# Impact of Data Normalization on Deep Neural Network for Time Series Forecasting

Samit Bhanja and Abhishek Das*, *Member, IEEE*

*Email- adas@aliah.ac.in

**Abstract**—For the last few years it has been observed that the Deep Neural Networks (DNNs) has achieved an excellent success in image classification, speech recognition. But DNNs are suffer great deal of challenges for time series forecasting because most of the time series data are nonlinear in nature and highly dynamic in behaviour. The time series forecasting has a great impact on our socio-economic environment. Hence, to deal with these challenges its need to be redefined the DNN model and keeping this in mind, data pre-processing, network architecture and network parameters are need to consider before feeding the data into DNN models. Data normalization is the basic data pre-processing technique form which learning is to be done. The effectiveness of time series forecasting is heavily depend on the data normalization technique. In this paper, different normalization methods are used on time series data before feeding the data into the DNN model and we try to find out the impact of each normalization technique on DNN to forecast the time series. Here the Deep Recurrent Neural Network (DRNN) is used to predict the closing index of Bombay Stock Exchange (BSE) and New York Stock Exchange (NYSE) by using BSE and NYSE time series data.

**Index Terms**— Neural Network, Deep Neural Network, Time Series, Data Normalization

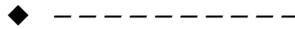

## 1 INTRODUCTION

FORECASTING the stock market price recently is gaining much more attention because if the successful prediction is achieved then the investors may be better guided. The profitability of the investment in stock market is highly dependable on prediction. Moreover the predicted trends also help the stock market regulator to make the corrective measure.

The Stock market is dynamic in nature and there are several complex factors that influence the market. Therefore the trends of the series are highly affected by these factors. Many researches have proposed many fundamental, technical or analytical models to predict the stock market[1]–[3] and give more or less exact prediction. There is some other linear approach such as moving average, exponential smoothening, time series regression etc. One of the most common and popular linear method is the Autoregressive integrated moving average (ARIMA)[4] model.

Artificial Neural Network (ANN) is an efficient computing system whose central theme is borrowed from the analogy of biological neural networks. The main objective of ANN[5] is to develop a system that can perform various computational tasks faster than the traditional systems. The ANN mimic the process of human's brain and solve the nonlinear problems, that's why it widely used for predicting and calculating the complicated task.

Nowadays, several Artificial Neural Network (ANN) models such as Multilayer Perceptron (MLP) neural network, Back Propagation (BP)[6][7] neural networks are used to predict the stock market price.

Artificial neural networks with back propagation learning algorithm[6] are widely used in solving various classification and prediction problems. Most of the shallow networks have one hidden layer, but Deep Neural Networks (DNN)[8]–[10] have many hidden layers, and these many hidden layer allow the user to build a network model to handle highly non-linear and dynamic data and functions.

Recently deep architectures are used in various fields such as object recognition, speech recognition, natural language processing, physiological affect modelling, etc., and they provide significant performance. Unfortunately, training deep architectures is a difficult task and the classical methods are highly effective when applied to shallow architectures, but they are inefficient when they are applied on the deep architectures.

The goal of this paper is to find out the effective data normalization method to predict the Indian stock market most efficiently.

In this paper, the Deep Recurrent Neural Network (DRNN)[8] is used as DNN and multi attribute stock market data as Time Series data[4]. The historical stock market data are applied on DRNN to train the DRNN. The DRNN acquire the knowledge by the training process and that knowledge is applied to predict the stock market.

The organization of the paper is as follows: Section 2 provides the preliminary concept of DNN. Section 3, represents the brief description of RNNs. In Section 4, description of time series data. Different normalization techniques are presented in Section 5. Data set description and proposed framework is represented in the Section 6. Section 7 describes the results and discussion and finally in Section 8, the conclusion is represented.

## 2 PRELIMINARY CONCEPT OF DEEP NEURAL NETWORK

The term deep neural network can have several meanings, but one of the most common is to describe a neural network that has two or more layers of hidden processing neurons. These multiple layers make the DNN so efficient and effective to process the non-linear and highly dynamic data. To process these types of dynamic information, so many effective and efficient deep learning algorithms have been developed. These deep architectures are composed of many layers of no-linear processing stages, where each lower layer's output is fed into the input of next higher layer.

Originally, the concept of deep learning was developed from ANN research. Hence the MLPs or Feed-forward neural network with many layers are the good example of DNN[8]. The Back-propagation (in 1980's) is a well-known algorithm for learning weights of these types of DNNs. But, in practical it is not alone suitable for training the networks that contain more hidden layers.

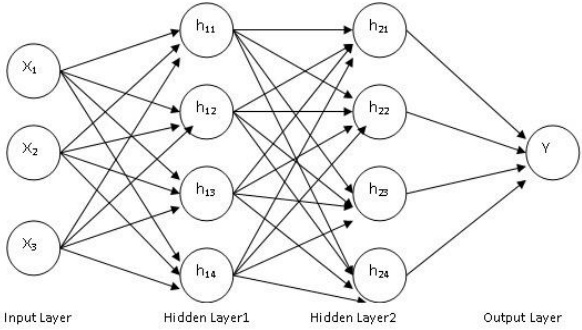

Fig. 1. A simple four layer DNN.

## 3 BRIEF DESCRIPTION OF RNNS

The RNNs[11] are one type deep generative architecture and they are often used to model and generate the sequential data. They are first introduced in 1986. RNNs are very powerful for modelling sequence data (e.g., speech or text).

RNNs are called recurrent because they perform the same task for every element of a sequence, with the output being dependent on the previous computations. Another way to think about RNNs is that they have a "memory"

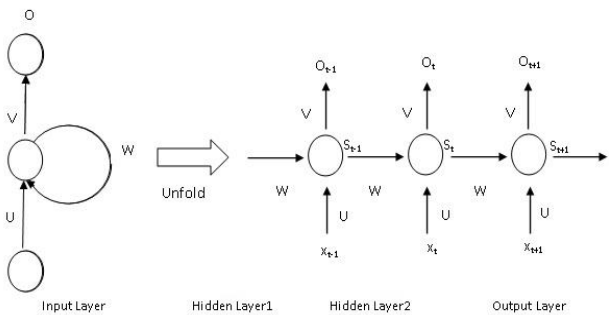

Fig.2. A recurrent neural network and the unfolding in time of the computation involved in its forward computation.

which captures information about what has been calculated so far. In theory RNNs can make use of information in arbitrarily long sequences, but in practice they are limited to looking back only a few steps. Here is what a typical RNN looks like:

The above diagram shows a RNN being unrolled (or unfolded) into a full network. The formulas that govern the computation happening in a RNN are as follows:

$x_t$ is the input at time step $t$.

$S_t$ is the hidden state at time step $t$. It's the "memory" of the network. $S_t$ is calculated based on the previous hidden state and the input at the current step: $S_t = f(Ux_t + WS_{t-1})$. The function $f$ usually is nonlinearity such as $tanh$ or $ReLU$.

$O_t$ is the output at step $t$. $O_t = softmax(VS_t)$.

Unlike a traditional deep neural network, which uses different parameters at each layer, an RNN shares the same parameters (U, V, W above) across all steps. This reflects the fact that the network performs the same task at each step, just with different inputs. This greatly reduces the total number of parameters are needed to learn.

## 4 DESCRIPTION OF TIME SERIES DATA

A time series is a series of data collected in equal time intervals such hour, day, month, year etc. This large volume of time series data are analysed for the forecasting, such as weather forecasting, stock market prediction, Sales forecasting etc[12]–[14]. Here it is expected that value of a data X in a given time is related to the previous values. The time series is measured for a fixed units of time, so the series of values can be expressed as

$X = \{x(1), x(2), \ldots, x(t)\}$, where $x(t)$ is the most recent value. For most of the time series problems, the goal is to forecast $x(t + 1)$ from previous values of the feature, where these values are directly related to the predicted value.

## 5 DIFFERENT NORMALIZATION TECHNIQUES

The effectiveness of any learning algorithm is heavily dependent on the normalization method[15]. The main objective of the data normalization method to produce a high quality of data that can feed into any learning algorithm. The time series data can have the wide range of values, so it need to be scale to a same range of values to speed up the learning process. There are so many data normalization techniques are available. Some of them are as follows –

### 5.1 Min-Max Normalization

In this approach, the data scaled to a range of [0, 1] or [-1, 1]. This method convert the a input value $x$ of the attribute $X$ to $x_{norm}$ the range$[low, high]$, by using the formula-

$$x_{norm} = \frac{(high - low) * (x - minX)}{maxX - minX} \qquad (1)$$

Where $minX$ and $maxX$ are the minimum and maximum value of the attribute $X$ of the input data set.

### 5.2 Decimal Scaling Normalization

In this technique, the decimal point of the values of the attribute is moved. This movement of the decimal point is depends on the number of digits present in the maximum value of all the values of the attribute. Hence a value $x$ of the attribute $X$ is converted to $x_{norm}$ by using the formula-

$$x_{norm} = \frac{x}{10^d} \quad (2)$$

Where $d$ is the smallest integer such that $Max(|x_{norm}|) < 1$

### 5.3 Z-Score Normalization

Z-Score or Standard Score converts all the values of an attribute to a common range of 0 and the standard deviation of the attribute. In this method the value $x$ of the attribute $X$ is converted to $x_{norm}$ by using the formula-

$$x_{norm} = \frac{x - \mu(X)}{\delta(X)} \quad (3)$$

Where $\mu(X)$ is the mean value and $\delta(X)$ is the standard deviation of the attribute $X$.

### 5.4 Median Normalization

In this method each input value of an attribute is normalized by the median of all the values of that attribute. By using the following formula, the normalized value $x_{norm}$ of the value $x$ of the attribute $X$ can be calculated.

$$x_{norm} = \frac{x}{median(X)} \quad (4)$$

### 5.5 Sigmoid Normalization

Here sigmoid function is used for the data normalization. The value $x$ othe the attribute $X$ can be normalized by the following sigmoid function-

$$x_{norm} = \frac{1}{1 - e^{-x}} \quad (5)$$

### 5.6 Tanh estimators

This method is one of the most powerful and efficient normalization technique. It is introduced by Hample. The normalized value $x_{norm}$ of the value $x$ of $X$ attribute of the input data set can be calculated by the following formula-

$$x_{norm} = 0.5 \left[ tanh\left[ \frac{0.01(x - \mu)}{\delta} \right] + 1 \right] \quad (6)$$

Where μ and δ are mean value and standard deviation of all the values of the $X$ attribute respectively.

## 6 DATA SET DESCRIPTION AND PROPOSED FRAME WORK

As the time series data, here the stock market data are taken for the prediction. The stock market data are nonlinear and highly dynamic in nature, so it is a very challenging task to successfully predict the market hypothesis.

In this work, the Bombay Stock Exchange (BSE)[16] and New York Stock Exchange (NYSE)[17] time series data are used for the training, validation and testing purpose. The basic indices such as opening price, high price and low price are taken for the input and closing price the output.

In this work, the BSE data from 1st January 2016 to 31st December 2017 (total 493 data points) are collected. Out of these 493 data 70% of labelled data (345 data points) is used for the training, 15% of labelled data (74 data points) is used for the validation and remaining 15% unlabelled data (74 data points) are used for the testing purpose. The NYSE data from 1st January 2016 to 31st December 2017 (total 503 data points) are also collected. Out of these 503 data points 70% of labelled data (352 data points) for training purpose, 15% of labelled data (76 data points) for validation purpose and remaining 15% of data (75 data points) are used for the testing purpose.

The architecture of the proposed model depicted in the figure 3. In this model, BSE and NYSE stock market data are collected. Before applying these data into the forecasting model, the data pre-processing is done. In pre-processing phase different normalization techniques are applied to scale the data into a certain range. Then these normalized data set is used in the forecasting model for the training, validation and testing purpose. In this work, deep recurrent neural network is used as the forecasting model.

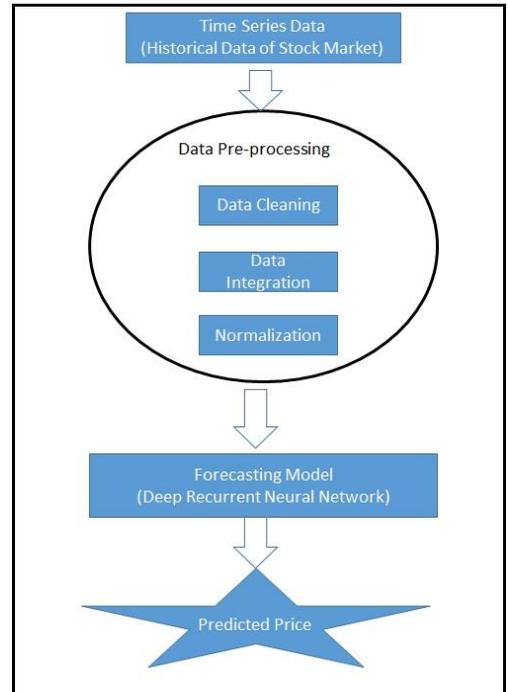

Fig.3. System Architecture

## 7 RESULT AND DISCUSSION

In this work, the stock indices like opening price, low price and high price are used for the input and closing price is used for the output of the network. The experiment is done using the Matlab software. The data are pre-processed by the MinMax, Decimal scaling, Z-Score, Me-dian, Sigmoid and Tanh Estimator normalization techniques. As the forecasting model here Deep Recurrent Neural Network (DRNN) with one input layer, 20 hidden layer and output layer is used. The performance of DRNN is measured for the different normalization technique with respect to the Mean Squared Error (MSE) and Mean Absolute Error (MAE).

Fig. 4 and Fig 5 show the graph of the predicted value of the closing price versus the actual value of BSE and NYSE stock market for the different normalization techniques respectively.

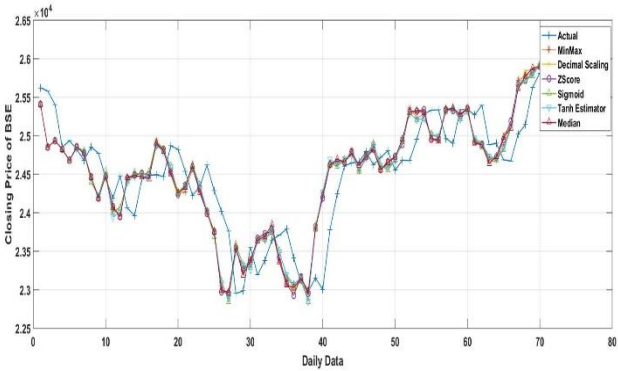

Fig. 4. Forecasting of closing price of BSE

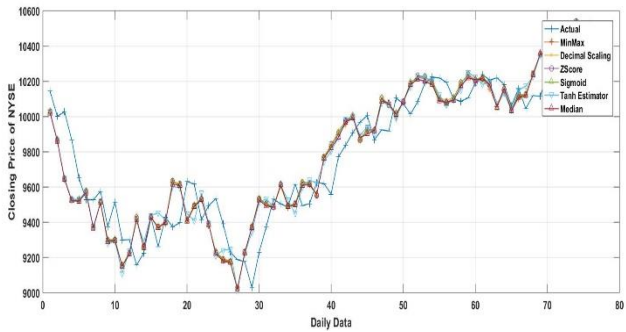

Fig. 5. Forecasting of closing price of NYSE

TABLE 1
PREDICTION ERROR OF BSE DATA FOR THE DIFFERENT NORMALIZATION METHODS

| BSE Data Set | | |
|---|---|---|
| Normalization Techniques | Prediction Errors | |
| | MSE | MAE |
| MinMax | 2.9079e-05 | 0.0040 |
| Decimal Scaling | 1.0823e-07 | 2.4352e-04 |
| ZScore | 3.0020e-04 | 0.0130 |
| Sigmoid | 2.9824e-08 | 1.2972e-04 |
| Tanh Estimator | 1.1537e-08 | 8.1233e-05 |
| Median | 2.1359e-06 | 0.0011 |

TABLE 2
PREDICTION ERROR OF NYSE DATA FOR THE DIFFERENT NORMALIZATION METHODS

| BSE Data Set | | |
|---|---|---|
| Normalization Techniques | Prediction Errors | |
| | MSE | MAE |
| MinMax | 1.9097e-05 | 0.0033 |
| Decimal Scaling | 1.6369e-06 | 9.5705e-04 |
| ZScore | 2.2066e-04 | 0.0111 |
| Sigmoid | 7.7942e-08 | 2.0090e-04 |
| Tanh Estimator | 1.5486e-08 | 9.2140e-05 |
| Median | 1.3853e-06 | 8.6526e-04 |

The Table 1 and Table2 shows the prediction errors such as MSE and MAE for the different normalization process for the BSE and NYSE stock market data respectively.

## 8 CONCLUSION

In this paper, the DRNN is applied to predict the sock index of BSE and NYSE stock market for the different normalization methods.

From the Fig.4. and Fig.5., it is observed that all the normalization techniques produced fair results.

Also, from the Table 1 and Table 2, it can be observed that all the normalization techniques produced very small amount prediction errors, but the Tanh Estimator normalization produce lesser MSE and MAE.

Thus it can be concluded that the Tanh Estimator normalization technique is an effective normalization method for time series prediction of Deep Recurrent Neural Network.

## ACKNOWLEDGEMENT

The Authors will like to thank Hewlett Packard (HP) for their generous grant of a HP Workstation Z4 G4 Pike along with two units of 24' Display devices to the Principal Investigator for conducting the research.

**Mr. Samit Bhanja** is a Ph.D scholar with the Dept. of Computer Sc. & Engineering at Aliah University, Kolkata and also working as an Assistant Professor, Dept. of Computer Science, Government General Degree College, Hooghly. He has a MTech in Computer Science and Engineering.

**Dr. Abhishek Das** is currently working as the Head and Associate Professor in the Dept. of Computer Sc. & Engineering at Aliah University, Kolkata. His Research Area is in Medical Image Processing, Computer Vision and IoT. He has received a Post Doctoral Research position from University of West Scotland, UK by a Fellowship received from the European Commission. He has also worked as an Assistant Professor in the Dept. of Information Technology, Tripura University ( A Central University), India. He has worked previously as a Reader in Computer Sc. & Engineering, Indian Institute of Space Science & Technology (IIST) and Lecturer in Information Technology, Bengal Engineering and Science University, Shibpur (now IIEST Shibpur) His Ph.D is from the Dept. of Computer Sc. & Engineering, Jadavpur University, India. His M.S. in Electrical & Computer Engineering from Kansas State University, USA and B.Tech in Computer Science & Engineering from Kalyani University, India. He has worked as a Business System Analyst in Blue Cross Blue Shield New York, USA & as a Quality Analyst in Vensoft Inc. Phoenix, USA and also as a Consultant Project Manager in the Software Industry. He is also a Visiting Scientist at Indian Statistical Institute, Kolkata. He has also worked in Technical Educational Administration as Regional Officer and Asst. Director (on deputation) at AICTE(under MHRD, Govt. of India). He is a NCERT National Scholar, New Delhi and a Tilford Dow Scholar, USA. His Biography has been published in the 28th edition of National Dean's List, USA. He is a Member of IEEE and a Honorary Senior Member of IACSIT Singapore. He has several publications in International and National Journals and peer-reviewed Conferences. He is also a TPC Committee Member of IEEE International conferences in South-East Asia and Editor of the Journal of Image Processing and Pattern Recognition progress(A UGC recognised Journal) and Special Issues Journal on Image Processing & Computer Vision of IGI Global Publishers, USA ( SCI and Scopus Indexed).